\documentclass[conference,a4paper]{IEEEtran}

\IEEEoverridecommandlockouts

\usepackage[hidelinks]{hyperref}
\usepackage[cmex10]{amsmath}
\usepackage{amssymb,amsfonts}
\usepackage{dblfloatfix}

\usepackage[ruled,vlined]{algorithm2e}
\usepackage{graphicx}
\graphicspath{{Figures/}}

\usepackage{booktabs}
\usepackage{siunitx}
\usepackage[numbers,compress]{natbib}
\usepackage{texnames}
\usepackage{bm,bbm}
\usepackage{orcidlink}

\usepackage{hyperref}
\usepackage[nameinlink]{cleveref}
\usepackage{tikz}

\usepackage[a4paper, total={7in, 9in}]{geometry} 
\usepackage{multirow}
\usepackage[skip=5pt]{caption} 
\begin{document}
\title{\uppercase{Ecological Mapping with Geospatial Foundation Models
}}

\author{
    \IEEEauthorblockN{
        Craig Mahlasi\textsuperscript{1}\orcidlink{0009-0002-5507-7358}, 
        Gciniwe S. Baloyi\textsuperscript{1}\orcidlink{0000-0002-8926-5595}, 
        Zaheed Gaffoor\textsuperscript{1}\orcidlink{0000-0003-1631-6627}, 
        Levente Klein\textsuperscript{1}\orcidlink{0000-0001-9497-1403}, 
        Anne Jones\textsuperscript{1}\orcidlink{0000-0003-1273-5533}, 
        Etienne Vos\textsuperscript{1}\orcidlink{0000-0003-2513-4980}, \\
        Michal Muszynski\textsuperscript{2}\orcidlink{0000-0002-0763-2612}, 
        Geoffrey Dawson\textsuperscript{1}\orcidlink{0000-0001-9873-2266},
        and Campbell Watson\textsuperscript{1}\orcidlink{0000-0003-3029-9069}
    }
    \IEEEauthorblockA{\textit{\textsuperscript{1}IBM Research}; \textsuperscript{2}\textit{IBM Sustainability Software}}
}

\maketitle

\begin{abstract}
The value of Earth observation foundation models for high-impact ecological applications remains insufficiently characterized. This study is one of the first to systematically evaluate the performance, limitations and practical considerations across three common ecological use cases: forest functional trait estimation, land use and land cover mapping and peatland detection. We fine-tune two pretrained models (Prithvi-EO-2.0 and TerraMind) and benchmark them against a ResNet-101 baseline using datasets collected from open sources. Across all tasks, Prithvi-EO-2.0 and TerraMind consistently outperform the ResNet baseline, demonstrating improved generalization and transfer across ecological domains. TerraMind marginally exceeds Prithvi-EO-2.0 in unimodal settings and shows substantial gains when additional modalities are incorporated. However, performance is sensitive to divergence between downstream inputs and pretraining modalities, underscoring the need for careful dataset alignment. Results also indicate that higher-resolution inputs and more accurate pixel-level labels remain critical for capturing fine-scale ecological dynamics.
\end{abstract}


\section{Introduction}

Geospatial foundation models (GFMs) are a fast-emerging area within the field of artificial intelligence (AI) models. The aim of these models is to learn robust and generalizable representations of the physical environment that can be adapted to a wide range of downstream tasks and applications while reducing the amount of labelled data needed. Most notable examples of GFMs include \citep{szwarcman2025prithvi, jakubik2025terramind,cong2022satmae, soni2025earthdial, fuller2023croma, xiong2024neural, mendieta2023gfm, mall2023remote}. While these models have shown value in general purpose downstream applications including general land cover mapping, segmentation of spectrally distinct features and non-complex scenario object-detection, these applications can be understood as being trivial. Conversely, ecological applications of GFMs are often not straightforward. This is because the intricate characteristics of the Earth’s biodiversity are not directly measured by earth-observing satellites or other data used in the pretraining of GFMs \citep{cavender2022integrating}. Even with fine-tuning, several limitations including low domain generalization, geographic bias and temporal bias have been observed \citep{mai2023opportunities,yang2025survey, xiao2025foundation}. For these reasons, task-specific machine learning supervised models remain a mainstay for ecological and conservation applications. 

In this paper, we explore the utility of GFMs for ecological mapping tasks. We do not seek to conduct a survey of GFMs, their capabilities and advances, for this we refer readers to \citep{mai2023opportunities, yang2025survey, xiao2025foundation, zhang2024towards, mendieta2023towards, ghamisi2025geospatial, janowicz2025geofm}. We focus on the utility, challenges and opportunities associated with developing high fidelity models for ecological applications using GFMs. To this end, we start by considering zero-shot land use land cover (LULC) mapping performance using a generative GFM. Secondly, we assess the performance of GFMs on specialized downstream tasks including forest trait mapping and peatland detection. Finally, we consider input data distribution shifts, label quality as well as the practicalities of multimodal foundation models.  
 
\section{Methods}

\subsection{Study sites and use cases}

\subsubsection{NEON sites}\label{NEON-sites}

The National Science Foundation's National Ecological Observatory Network (NEON) sites are a network of ecologically important field sites located across the United States \cite{neon-link}. Long-term ecological data are collected from these sites, including characteristics of plant and animal life, soil, nutrients, freshwater and atmospheric dynamics, with the aim of better understanding the impact of climate change on biodiversity. Here we explore the usefulness of GFMs performing forest trait mapping across the NEON sites, focusing on leaf form and canopy cover density segmentation. 

\subsubsection{Karukinka Nature Park}

Karukinka Nature Park, located in southern Patagonia on the archipelago of Tierra del Fuego, is a world-renowned ecological hotspot \cite{karikinka-link}. This region contains extensive and well developed peat bogs, which contribute significantly to global carbon sequestration \cite{vanBellenetal2016}. The formation of the peatlands is driven by the presence of Sphagnum Magellanicum, a peat moss common in the region. From an aerial perspective, this species of peat moss commonly appears reddish-brown coloured. Here we explore the utility of GFMs to delineate and monitor peatland extent.

\subsection{Models}

In this work, we use two GFMs and one pretrained deep learning model: Prithvi-EO-2.0 \citep{szwarcman2025prithvi}, TerraMind \citep{jakubik2025terramind}, and ResNet-101 \citep{he2016deep}. Both Prithvi and TerraMind are transformer-based GFMs developed specifically for and pretrained using Earth Observation data, whereas ResNet is a convolution based model pretrained on RGB images collected online and spanning thousands of categories. TerraMind is designed to operate over and generate multimodal Earth Observation representations, enabling joint reasoning across heterogeneous geospatial inputs. We consider ResNet a strong baseline due to its inductive bias and robust performance in limited-data regimes. Model training and fine-tuning were performed using TerraTorch, a toolkit built for GFM fine-tuning and benchmarking \cite{terratorch}. TerraTorch is built on PyTorch Lightning and therefore inherits PyTorch Lightning's training, validation, and testing framework through low-code YAML configuration files. 

\subsection{Input modalities}

For each NEON site, Copernicus Sentinel-2 L2A (S2-L2A) \cite{s2-l2a} cloud free composites were generated for the summer months (May - August) of 2019. While for the Karukinka study area, S2-L2A cloud free composites were generated for the summer months (October to March) of 2018-2019. For the multimodal experiments, the RGB and NDVI modalities where computed from the S2-L2A cloud free composites. For experiments that include Copernicus Sentinel-1 GRD (S1-GRD) \cite{s1-grd} modality, monthly mosaics for IW variant in the VV and VH decibel gamma bands for January - March 2019 where collected. Thereafter the mean of VV and VH bands where computed across the three months. For the DEM modality, the Copernicus GLO-30 Digital Surface Model where collected for the various study sites \cite{glo-30}. The PEATGRIDS-NDVI approach utilised S2-L2A cloud free composites, which were computed for each of the five global peatland locations. All modalities were regrid and aligned to the resolution (10m) and coordinate reference system of S2-L2A dataset. Finally, the data (inputs and labels) were split into patches of size 224 x 224 pixels. 

\subsection{Deriving labels }

\subsubsection{Forest traits mapping}

The Copernicus Land Monitoring Service Global Dynamic Land Cover data provide yearly global estimates of LULC across 23 classes from 2015-2019 at 100m resolution \cite{cop-2019}. This LULC dataset contains 12 forest categories that classify forests based on three criteria, namely canopy density, leaf form, and leaf retention. For leaf form segmentation, the forest categories were extracted and merged into 5 classes: needle leaf forests, broad leaf forests, mixed leaf forests, unknown leaf form forests, and other (non-forested). While for canopy density, the 12 forest classes where merged into three broad categories, classifying open forest (canopy cover is between 15-70\%), closed forest (canopy cover is $>$70\%), and other (non-forested). 

\subsubsection{Peatlands detection} \label{peatland-labels}

Several datasets were adapted to derive a total of three unique label datasets that roughly define the extent of the peatlands in the study area. The first dataset is the Copernicus Land Monitoring Service Global Dynamic Land Cover for 2020 \cite{cop-2020} (henceforth, this shall be known as COP 2020 LULC) which provides classifications for 11 land cover classes at a 10m resolution. Here, the herbaceous vegetation and herbaceous wetland classes are extracted, which appear to overlap well with the presence of peat moss. Thereafter, to eliminate false labels on mountain tops and steeply dipping slopes, all labels above an elevation of 450m and a slope angle above 10\textdegree, were removed. The second dataset is the Chilean National Forestry Corporation (CONAF) inventory of native vegetation resources dataset which classifies vegetation land cover for the entire Chile \cite{chilean-lulc}. Here, we extract the peat bog class (turbales in Spanish) as labels. The third and final dataset is the PEATGRIDS dataset which contains peat thickness and carbon stock maps estimated over peatland areas across the globe at 1$\times$1km resolution \cite{peatgrids}. The PEATGRIDS data is sampled at five geographically distinct peatlands across the world: Karukinka, Chile; Ontario, Canada; Luginetskiy, Russia; Zhytomyr Oblast, Ukraine; Louisiana, USA. It is further processed with NDVI computed from the S2-L2A cloud free composites at each location. A threshold of 0.23 - 0.8 \cite{anderson-ndvi,chavez-ndvi} is applied to the NDVI layers to produce a binary mask classifying peat from non-peat, and the labels defined at the intersection of the PEATGRIDS positive labels and the NDVI positive labels. 

\subsection{Experiments}

\subsubsection{TerraMind generation}

The mapping of ecological processes often requires timely updates of data (such as earth observation data) and the outputs derived from the said data. However, sub-optimal acquisition conditions and long revisit times often hinder the application of earth observation data in ecological mapping. With the TerraMind model, however, this data limitation can be partially addressed through the "any-to-any" modality generation feature, where any of the input modes can be used to generate or translate into any of the other modes, thereby, filling in the data gaps. Here, we test TerraMind's zero-shot capability in generating ESRI LULC modality based on the S2-L2A modality and evaluate it against the original ESRI global 10-class land cover dataset \cite{esri-lulc}. 

\subsubsection{Forest traits mapping} 

To assess the utility of GFMs on forest traits mapping, two tasks were chosen: leaf form and canopy cover density segmentation. The experiments focus on the NEON sites described in section \ref{NEON-sites}. S2-L2A cloud free composites were used as inputs. We fine-tune Prithvi, TerraMind, and ResNet on a single A100-80GB GPU. The AdamW optimizer was used with a learning rate of $1\times10^{-4}$, the dice loss function and a weight decay configuration of $5\times10^{-2}$ were specified. 

\subsubsection{Peatland detection} 

Here, we fine-tune TerraMind, Prithvi and ResNet models with all three label datasets (section \ref{peatland-labels}) to segment the peatlands from the surrounding land cover types. For Prithvi and ResNet, only unimodal fine-tuning experiments where conducted using 12 S2-L2A bands. For TerraMind, both unimodal (S2-L2A) and multimodal (S1-GRD, S2-L2A, RGB, NDVI and DEM) fine-tuning experiments where conducted. All fine-tuning experiment where run on single A100-80GB GPU where the AdamW optimizer was used with a learning rate of $1\times10^{-4}$ and the dice loss function and a weight decay configuration of $5\times10^{-2}$. 

\section{Results and Discussion}
\label{sec:results}

\subsection{Generation}

\begin{figure}[hbt!]
    \centering
    \includegraphics[width=\linewidth]{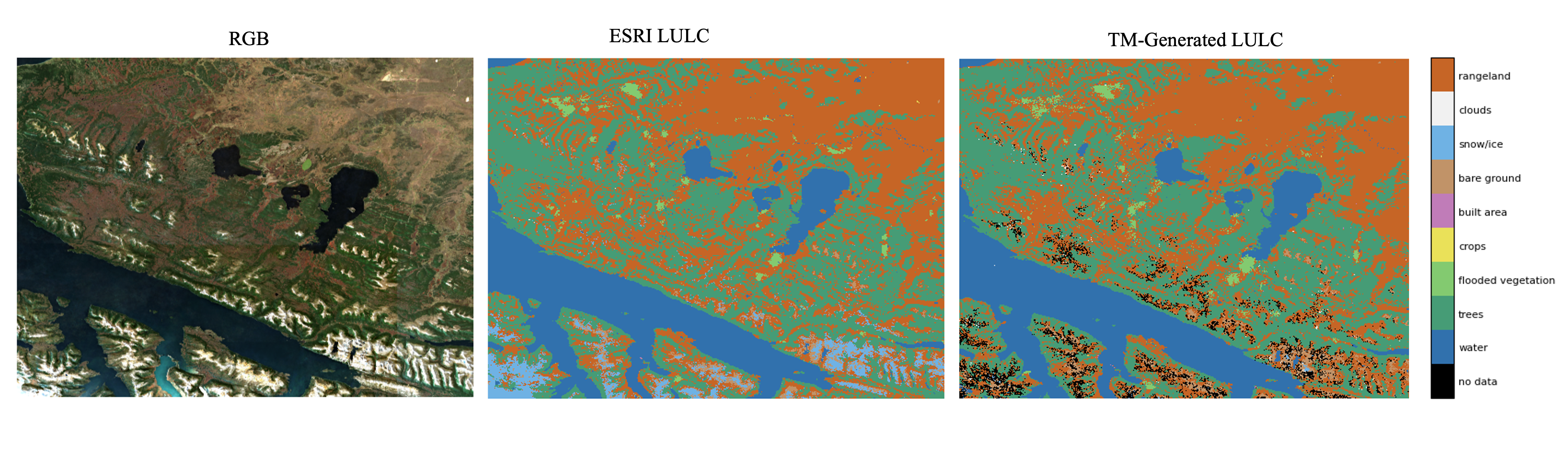}
    \caption{Generated LULC map of the Karukinka region using TerraMind.}
    \label{fig:gen}
\end{figure}

Having started with S2-L2A data as input to the TerraMind model, Fig. \ref{fig:gen} shows the qualitative evaluation of the generation output, whilst Table \ref{tab:gen} shows the quantitative performance for ESRI LULC generation for the Karukinka study area. Overall, the model achieved a weighted intersection over union (IoU) score of 78.82\%, which indicates strong agreement between the LULC classes generated by the model and the ESRI ground truth that we were comparing it against. Moreover, the model performed strongly in generating vegetation classes, namely trees (90\%), and rangelands (87\%). Bare ground and flooded vegetation came in fourth and fifth with F1-scores of 50\% and 48\% respectively. In respect to Figure \ref{fig:gen}, the rangeland class overlaps well with the presence of peatlands, but struggles to separate from other herbaceous vegetation. 

\begin{table}[hbt!]
\caption{Performance evaluation from TerraMind generation.}
\label{tab:gen}
\centering
\renewcommand{\arraystretch}{1.1}
\begin{tabular}{lll}
\hline
\textbf{Classes} & \textbf{Support} &  \textbf{F1-score}\\ \hline 
Water             & 13743498      & 0.99            \\ 
Trees             & 31735327      & 0.90          \\ 
Flooded vegetation  & 826811        & 0.50           \\ 
Crops             & 3140          & 0             \\ 
Built area        & 2231          & 0.13          \\ 
Bare ground       & 1243421       & 0.48          \\ 
Snow/ice          & 2612880       & 0.02         \\ 
Clouds            & 5002          & 0          \\ 
Rangeland         & 33721962      & 0.87         \\ 
\hline
\end{tabular}
\end{table}

\vspace*{-5mm}
\subsection{Forest traits mapping}
\label{sec:FM results}

\begin{figure}[hbt!]
    \centering
    \includegraphics[width=0.48\textwidth]{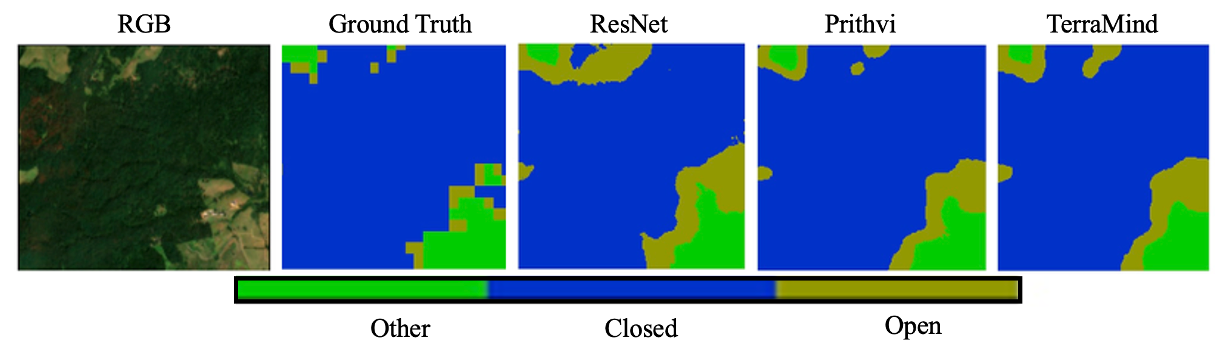}
    \caption{Visual comparison of canopy cover density prediction by ResNet, Prithvi and TerraMind.}
    \label{fig:ccd}
\end{figure}

\begin{figure}[hbt!]
    \centering
    \includegraphics[width=0.48\textwidth]{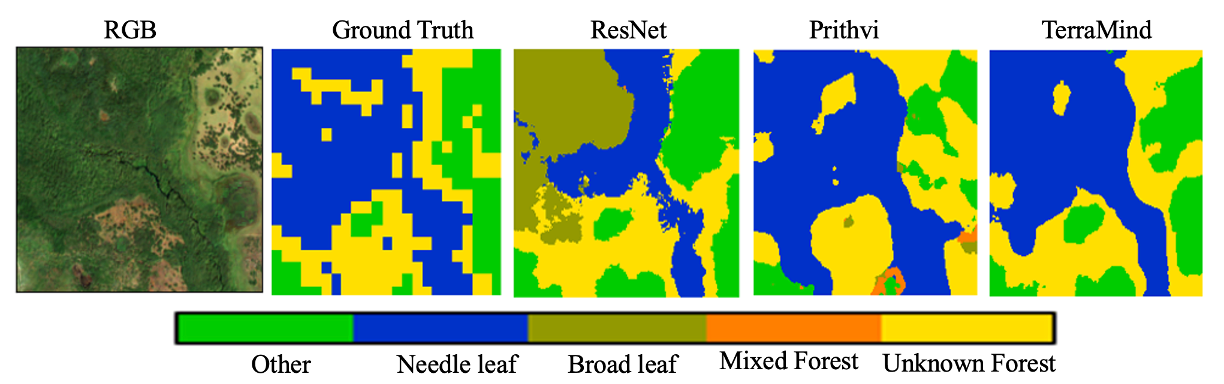}
    \caption{Visual comparison of Leaf form prediction by ResNet, Prithvi and TerraMind. }
    \label{fig:lf}
\end{figure}

The results shown in Table \ref{tab:FFM1} and \ref{tab:FFM2}, indicate superior performance of GFMs for both canopy density and leaf mapping, with both out performing ResNet by at least 20\%. TerraMind was the best performing model, followed closely by Prithvi in both experiments. Fig \ref{fig:ccd} and \ref{fig:lf} are qualitative evaluations of the canopy cover density and leaf form segmentation tasks respectively. The figures illustrate how Prithvi and TerraMind outperform ResNet in both segmentation tasks. We can see that while the models where fined-tuned with coarse-grained targets, they were able to capture the spatio-spectral dimensions of the inputs and targets. We note that ViT based models (Prithvi and TerraMind) outperformed the ResNet. 

\begin{table}[hbt!]
\caption{Performance evaluation of ResNet, Prithvi and TerraMind on canopy cover density mapping.}
\label{tab:FFM1}
\centering
\renewcommand{\arraystretch}{1.1}
\begin{tabular}{lcccc}
\hline
\textbf{Model} & \textbf{F1-score} & \multicolumn{3}{c}{\textbf{IoU per  Class}} \\ 
\cline{3-5}
               &                   & \textbf{Other} & \textbf{Open} & \textbf{Closed} \\ 
\hline
ResNet          & 0.54              & 0.38             & 0.47             & 0.17             \\ 
Prithvi        & 0.74              & 0.58             & 0.68             & 0.40            \\ 
TerraMind      & 0.75              & 0.58             & 0.69             & 0.41             \\ 
\hline
\end{tabular}
\end{table}

\begin{table}[hbt!]
\caption{Performance evaluation of ResNet, Prithvi, and TerraMind on leaf form mapping.}
\label{tab:FFM2}
\centering
\resizebox{0.5\textwidth}{!}{%
\fontsize{52}{54}\selectfont
\begin{tabular}{lcccccc}
\hline
\textbf{Model} & \textbf{F1-score} & \multicolumn{5}{c}{\textbf{IoU per class}} \\
\cline{3-7}
 &  & \textbf{Other} & \textbf{Needle leaf} & \textbf{Broad leaf} & \textbf{Mixed forests} & \textbf{Unknown forests} \\
\hline
ResNet     & 0.41 & 0.42 & 0.21 & 0.17 & 0.03 & 0.18 \\
Prithvi   & 0.67 & 0.56 & 0.43 & 0.56 & 0.33 & 0.42 \\
TerraMind & 0.70 & 0.63 & 0.55 & 0.52 & 0.42 & 0.43 \\
\hline
\end{tabular}}
\end{table}

\vspace*{-5mm}
\subsection{Peatland detection}

\begin{figure}[h!]
    \centering
    \includegraphics[width=0.48\textwidth]{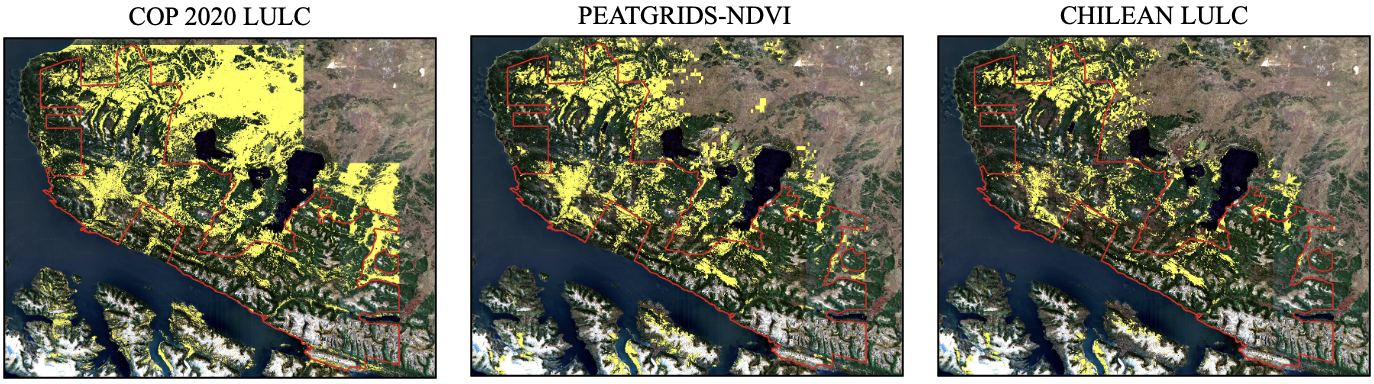}
    \caption{Comparison of the three labels datasets derived for peatlands detection. Labels shown in yellow.}
    \label{fig:peatlands-labels}
\end{figure}

Figure \ref{fig:peatlands-labels} illustrates the three labels datasets derived in section \ref{peatland-labels} for fine-tuning of the GFMs. The COP 2020 LULC labels, overlap well with the presence of the peat moss. However, none of the input modalities except the RGB, can effectively distinguish between different subclasses of vegetation, such as distinguishing between peat moss and other herbaceous vegetation. This is even more pronounced for the Chilean LULC labels, where only certain red-brownish patches (signifying the presence of peat moss) are labelled as peatlands. The PEATGRIDS-NDVI labels align well with the presence of peat moss, and avoid labelling the north east section as peats. However, the coarseness of the original PEATGRIDS dataset, introduces artefacts that prevent the accurate delineation of boundaries. 

Table \ref{tab:peatlands-detection-results} illustrates the performance of the three models. For the unimodal experiments, Prithvi and TerraMind outperform the ResNet model across all label datasets. The multimodal experiments with TerraMind produce superior performance in separating the peats from non-peats. Perhaps owing to the inclusion of S1, NDVI and DEM, which may help distinguish peatlands from other vegetation classes. Nonetheless, the Chilean LULC labels are the most challenging to reproduce.

\begin{table}[hbt!]
\caption{Performance evaluation of ResNet, Prithvi and TerraMind on peatland detection.}
\label{tab:peatlands-detection-results}
\centering
\renewcommand{\arraystretch}{1.1}
\begin{tabular}{llccc}
\toprule[.1em]

\textbf{Model}  &\textbf{Labels}& \textbf{F1-score} & \multicolumn{2}{c}{\textbf{IoU per  Class}}\\
\midrule[.1em]
&&& \textbf{Other}& \textbf{Peatland}\\     
\midrule[.1em]

\multirow{4}{*}{ResNet} 
&COP 2020 LULC& 0.86& 0.84& 0.44\\
\addlinespace
& PEATGRIDS-NDVI& 0.59& 0.66&0.10\\
\addlinespace
& CHILEAN-LULC& 0.90& 0.89&0.10\\
\midrule[.1em]

\multirow{4}{*}{Prithvi}
&COP 2020 LULC& 0.90& 0.88& 0.60\\
\addlinespace
& PEATGRIDS-NDVI& 0.81& 0.81&0.59\\
\addlinespace
& CHILEAN-LULC& 0.91& 0.91&0.17\\
\midrule[.1em]

\multirow{4}{*}{TerraMind}       
&COP 2020 LULC& 0.89& 0.87& 0.57\\
\addlinespace
& PEATGRIDS-NDVI& 0.88& 0.84&0.67\\
\addlinespace
& CHILEAN-LULC& 0.92& 0.92&0.19\\
\midrule[.1em]

\multirow{3}{1cm}{TerraMind (multimodal)}
& COP 2020 LULC*& 0.90& 0.82&0.77\\
\addlinespace
& CHILEAN-LULC& 0.95& 0.95&0.34\\ 
\bottomrule[.1em]
\multicolumn{5}{l}{\small \textit{*Uses S2-L2A, NDVI, and DEM modality only.}} \\
\end{tabular}
\end{table}

\begin{figure}
    \centering
    \includegraphics[width=0.48\textwidth]{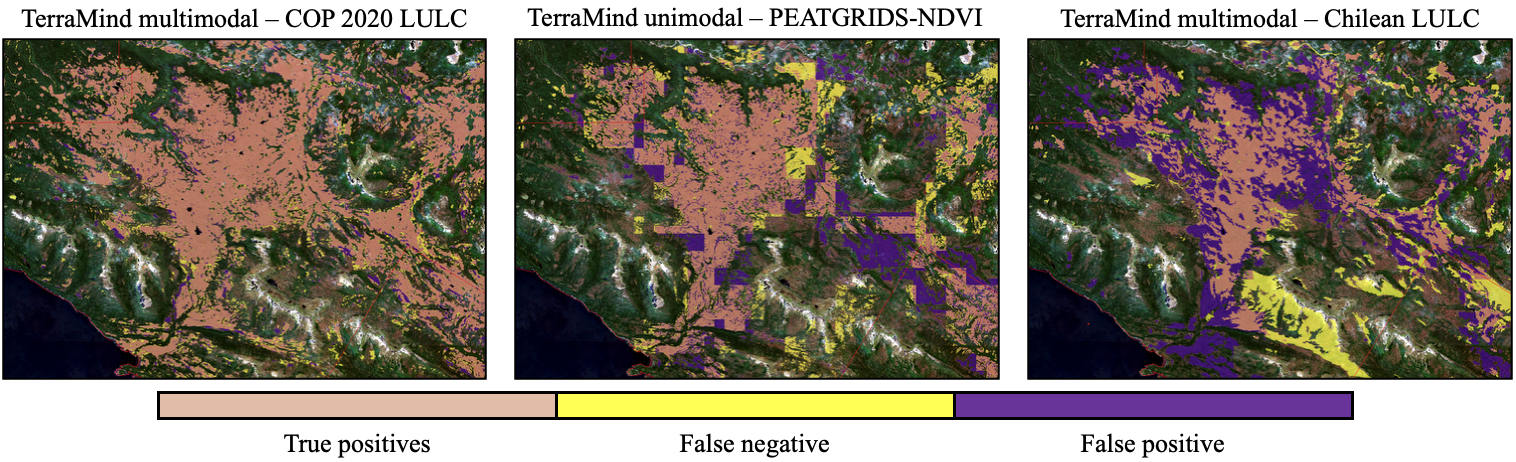}
    \caption{Error map of peatland predictions over groundtruth for three experiments}
    \label{fig:peatland-predictions}
\end{figure}

Lastly, Figure \ref{fig:peatland-predictions} illustrates the predictions for the most performant model for each label dataset. Assuming the presence of peat moss (reddish-brown patches) to be the intuitive target, models produced using the COP 2020 LULC produce a high degree of false positives. On the other hand, the PEATGRIDS-NDVI and Chilean LULC has fewer false positives and a high degree of false negatives. We attribute this to the fact that none of the models are pretrained on data that represent subsurface dynamics, such as the presence of a peat layer, carbon content, or subsurface hydrodynamics. 

\section{Conclusion}

The results for the experiments illustrate moderate performance for the various use cases, when compared to the general SOTA currently \cite{Saimun2025,Li31122024,ZHANG2024,CHENG2025100165}. While this is true across the board, it is clear that the GFMs produce superior performance compared to the baseline ResNet models. Pretraining GFMs on earth observation data is the primary value of GFMs, because they address the domain-gap which often is the cause of low performance when using non-geospatial backbones such as ResNet for geospatial applications. The marginally higher performance of TerraMind over Prithvi is attributed to the fact that Prithvi was pretrained on 6 HLS bands while TerraMind was pretrained on all 12 S2-L2A bands. However, with the addition of supplementary modalities, TerraMind significantly outperforms the unimodal counterparts.

Consideration should also be given to the quality of the input data, in the context of model performance. While not shown in this paper, the performance of the models vary significantly when using single date satellite acquisitions versus temporal aggregates. Temporal bias is also evident in model predictions. In-addition, the nominal resolution of the input data (10m) coupled with the inability of transformers to recover pixel-level details, reduces the ability to detect small scale features or changes in vegetation. In this case, the models may benefit from high resolution inputs. Furthermore, depending on the region (persistent cloud cover), the consistent use of either temporal aggregates or unprocessed single time acquisition, should be considered. Ideally, multitemporal sampling when creating fine-tuning datasets should be considered.  

From the label data perspective, the resolution and ambiguity of the labels, significantly affect model performance. For example, the COP 2019 labels used in the forest traits mapping, are roughly at a 100m, which do not accurately delineate the boundaries between classes. This reduces the validity of the model assessment. Caution should thus be given, when using "off-the-shelf" datasets as labels, especially when initially derived through machine learning techniques. While not always possible, preference should be given to label datasets derived from field based observation, and corroborated by experts. Here the models may benefit from high fidelity hand-picked labels.

\section*{Acknowledgment}

This work was supported by the MIT-IBM Watson AI Lab and Goldman Sachs.

\small
\bibliographystyle{IEEEtranN}
\bibliography{references}

\end{document}